# Big Models for Big Data using Multi objective averaged one dependence estimators

Mrutyunjaya Panda, Department of Computer science and Applications,

Utkal University, Vani Vihar-751004, India

mrutyunjaya74@gmail.com

**Abstract**

Even though, many researchers tried to explore the various possibilities on multi objective feature selection, still it is yet to be explored with best of its capabilities in data mining applications rather than going for developing new ones. In this paper, multi-objective evolutionary algorithm ENORA is used to select the features in a multi-class classification problem. The fusion of AnDE (averaged n-dependence estimators) with n=1, a variant of naive Bayes with efficient feature selection by ENORA is performed in order to obtain a fast hybrid classifier which can effectively learn from big data. This method aims at solving the problem of finding optimal feature subset from full data which at present still remains to be a difficult problem. The efficacy of the obtained classifier is extensively evaluated with a range of most popular 21 real world dataset, ranging from small to big. The results obtained are encouraging in terms of time, Root mean square error, zero-one loss and classification accuracy.

1. **Introduction**

Feature subset selection basically aims at providing the best possible feature subsets out of the total features available in the dataset in order to reduce the computational overhead in learning, leading to build an simple yet an accurate classifier. It is also noticed that, there are a plenty of approaches present in the literature for obtaining quality feature subsets, still it is considered to be a difficult task to obtain the best ones [1]. In order to address such complex computational problems arising from the large size of the input data, researchers are inspired towards using nature inspired algorithm recently to perform better optimization in classification task and evolutionary algorithms are one of them [2]. The parallel architecture of the Evolutionary algorithms (EA) are a potential candidate to process such big data automatically for optimal parameter setting and more importantly obtaining a viable solution for better interpretation of the model with best classification accuracy possible [3,4].While in single-objective optimization criteria, a single solution exists for which all criteria are optimal; Considering feature subset selection process as a multi-objective one, there is no single optimal solution that can outweigh the other, hence a solution of Pareto-optimal set is obtained. The Pareto-optimal set solutions are considered not to be dominated by any other possible solution [5,6].

**Motivation**
In some cases the amount of features can make construction of an induction model hard, either because the model cannot fit in memory or construction would take too long. Creating a limited subset can help construct a model within these constraints.
It is observed that solving multi-objective optimization problems using the exact methods such as: linear programming and gradient search etc. is too complex a process that can easily be over shadowed by using evolutionary algorithms. This is because of the parallelism and finding similarities by recombination process by the multi-objective optimization process, that make it a interesting area of research in many diverse applications. In spite of all, still there is a lack of studies that compare the performance and different aspects of these approaches. This motivated us to understand and carry out research, whether the multi-objective optimization approach suits to different problems at hand and Consequently, able to build a effective and efficient classifier with small as well as big dataset.

The rest of the paper is organized as follows: Section 2 discusses some related research followed by concepts of feature selection methods and MOEA are described in Section 3. The AODE classifier is discussed in Section 4 and then, materials and methods used for the experimentation presented in Section 5. Section 6 provides the experimental results and discussion on the proposed approach. Finally, we conclude in Section 7 with future scope of research.

2. **Related work**

This section discusses the relevance and possible applications of evolutionary algorithms, particularly genetic algorithms, in the domain of knowledge discovery in databases by different researchers. Dehuri and Ghosh [7] proposes to use a multi-objective genetic based feature selection method for knowledge discovery process and discusses its pros and cons for its validity and potentiality. Ducange, Lazzerini and Marcelloni [8] used NSGA-II, a well known multi-objective optimization algorithm as a feature subset selection in highly imbalanced dataset. They perform the classification and presented their suitability with sensitivity, specificity and interpretability in ROC plane. Laura Emmanuella et al. [9] proposed to use ensemble filter based classification with feature subset selection done by Particle swarm (PSO), genetic algorithm (GA) and ant-colony optimization (ACO) techniques. They perform both mono-objective and bi-objective versions for feature selection and finally claimed that the PSO based ensemble classifiers with bi-objective version performs well in 11 datasets investigated. Khan and Baig [10] opined that large number of irrelevant attributes are to be removed from the datasets in order to enhance the accuracy of the classifier. They use latest multi-objective genetic algorithm (NSGA-II) for feature subset selection and then applied with reduced feature sets to ID3 decision tree classifier on several datasets collected from UCI machine learning repository. They finally conclude with a feasibility study with NSGA-II method with individual attributes of the datasets used for investigation. Cataniaa, Zanni-merka, Beuvrona and Colleta [11] discussed the various aspects of multi-objective optimization NSGA-II implementation in Patient Transport services with preliminary results. They compared the results with the itineraries proposed by human experts and found it satisfactory. The authors Stoean and Gorunescu [12] presented a study to investigate the requirement of evolutionary algorithms to relief the user from the burden of curse of dimensionality. They applied in medical diagnosis considering that feature ranking of the attributes are of paramount importance for better decision making. Abd-Alsabour [13] presented a good review on some of the most recent feature selection methods based on evolutionary algorithms. They discussed about the pros and cons of such a method with their theoretical **issues.** Mukhopadhyay, Maulik, Bandyopadhyay, and Coello Coello [14] presented a comprehensive survey on recent advances in multi-objective evolutionary optimization techniques as a feature selection and classification for automatic processing of large qualities of data that can solve many real world problems with various various conflicting measures of performance. Cannas, Dessi and Pes [15] proposed four popular filter based procedure along with support vector machine in order to reduce the search space in high dimensional micro array data sets, for obtaining potential solutions to predict and diagnose the disease. A generic review on various filter, wrapper based feature selection and their possible combination are presented by Chandrasekharan and Sahin [16]. They also stated that the beauty of such methods lead to fast, accurate and simple classifiers, after investing their applicability on several standard dataset. Chen and Yao [17] demonstrated the effectiveness of the proposed ensemble of evolutionary multi-objective algorithms and Bayesian automatic relevance determination to obtain better accuracy with reduced feature sets. They conclude that their proposed approach applied in several real world scenarios outperforms other available ensemble approaches. The concept of dominance is used as a part of multi-objective genetic algorithms by the authors Macintyre et al. [18] where they pointed out the tradeoffs between computational complexity with classification accuracy. In order to validate their results, they used neural network and neuro-fuzzy hybrid in two small and high dimensional regression problems. The authors Jim´enez et al. [19] presented a good research by stating the usefulness of multi-objective evolutionary algorithms in dealing with the cases when number of attributes are very high which is already proven to be the best for attribute selection. They applied ENORA to Multi-Skill Contact Center Data Classification as a posteriori process in a multi objective context and finally compared with NSGA-II for validation of their findings. Ariasa, Gameza, Nielsenb and Puertaa [20] proposed a scalable pair wise class interaction framework for multi-dimension classification by taking several base classifiers and inference methods afterwards. They perform their experiments on a wide range of publicly available dataset and conclude that their approach is efficient in comparison with other exiting straw-men methods. A novel multi-objective genetic algorithm based feature selection combined with support vector machine is used [21] for selection of genes in micro array dataset for cancer diagnosis. They conclude that their approach with two or three objectives based on sensitivity and specificity quality measures are highly appropriate in comparison to the other existing algorithm available in the literature. Shi, Suganthan and Deb [22] proposed to use support vector machine as a classifier for structural classification of protein (SCOP) after the relevant feature are selected by using Multi-Objective Feature Analysis and Selection Algorithm (MOFASA). They conclude with their promising results with its applicability to future biological analysis. Datta, Deb, Fonseca, Lobo and Condado [23] proposed a spatial GIS based multi-objective evolutionary algorithm (NSGAII-LUM) for better understanding of the impact from land uses in Mediterranean landscape from Southern Portugal.

Multi-objective optimization (MOO) task to SVM design is evaluated in the real-time detection of pedestrians in infrared images for driver assistance systems, as a pattern recognition task. They discuss about how to model the proposed framework by considering the tradeoffs between accuracy and model complexity with reduced number of support vector [24]. Tan, Teoh, Yua and

Goh [25] proposed genetic- support vector (GA-SVM) hybrid evolutionary algorithm for attribute selection in data mining. They concluded that their approach applied to real life data sets obtained from UCI machine repository is accurate and consistent, as compared to the several well established algorithms. The authors provided an empirical comparison of various MOO algorithms, considering six test functions, which are aimed at giving an impression to understand which technique performs well under what condition in Zitler, Deb and Thiele [26]. A good tutorial on theory and model are provided by Zitzler, Laumanns, and Bleuler [27], where various algorithmic aspects of MOO such as: assignment of fitness function, elitism and diversity etc. Are discussed. They addressed on how to simplify the MOO by method of exchange with many standard applications.

3. **Evolutionary Multi-objective Optimization (MOEA )**

Application of multi-objective optimization starts with a prior knowledge on search technique that will lead to get Pareto optima solutions vis-a-vis preventing premature convergence. It is also pointed by Zitzler et al. [27] that due to the complex nature of generating Pareto optimal solution sets, many stochastic search namely: evolutionary algorithms, simulated annealing etc. are developed in order to find a good approximation to the near optimal solutions. The beauty of the evolutionary search lies in combining the several solutions in a recombination process to obtain the new best solutions. Further, it can be noticed that MOO can be a maximization or minimization problem with a number of objective functions [28]. The basic steps involved in such a method used in feature selection process can be observed in Figure 1.
Now-a-days, multi-objective evolutionary algorithms have attracted many a researchers in the multi faced applications to provide some viable solutions to multi-objective problems [28-29]; that include to name a few: Non-dominated sorting Genetic algorithm (NSGA) by Deb and Srinivas [30]; Strength Pareto Evolutionary Algorithm (SPEA 1 and 2) [31-32].

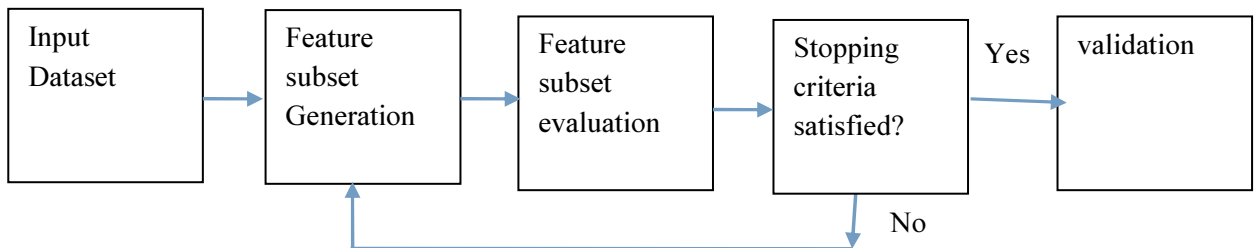

**Figure: 1 Feature selection process**

Any feature selection strategy starts with feature subset generation as a first step where some kind of search is done to obtain the candidate feature subset followed by candidate subset evaluation. In candidate subset evaluation, each candidate subset is compared and evaluated with previous best ones with some criteria, then the best one replaces the others. This process continues till stopping criteria is met and finally, the best selected candidate subsets are validated for their suitability on dataset under consideration.
One step further in this direction, one can think of using evolutionary algorithms to solve MOO problems, for their inherent capability to deal with set of candidate subset solutions or population. This makes it possible to find the Pareto optimal set in a single go rather than going for several runs as in the case of the traditional optimization techniques.

Multi-objective evolutionary algorithm based optimization (MOEA) mainly tries to address the following issues: (1) How best the Pareto-optimal set is obtained through proper assignment and selection of fitness functions and (2) prevention of convergence at an early stage by intelligently decide the diverse population.

The two most popular MOEA are ENORA (Elitist Pareto-based multi-objective evolutionary algorithm for diversity reinforcement) and NSGA-II (elitist non-dominated sorting genetic algorithm). They are discussed below.

### 3.1. ENORA

ENORA [33] is an elitist Pareto- based multi-objective evolutionary algorithm that uses a Mu plus Lambda survival with uniform random initialization and binary tournament selection process for exploring the attribute space. Crowding distance is used as a measure to find ranking on local non-domination level along with self adaptive uniform crossover and self adaptive single bit flip mutation. Ad-hoc elitist generational replacement technique is used to maintain diversity among the individuals.

In this paper, multi-objective evolutionary search is used as feature subset selection process using ENORA and the process seems to be maximization-minimization task. First, maximization process is set by the evaluator and then subset cardinality minimization is done as second step. Finally, the non-dominated Pareto optimal solution in the last population with best fitness is chosen as output.

To illustrate the process, for example: Given a population P of N individuals, N children are generated by random initialization and binary tournament selection, crossing and mutation. The new population is obtained by electing the N best individuals from the union of parents and children. The ranking of individuals are done by the operator best where each individual comes with a ranking in its slot and their corresponding crowding distance. One individual (x) is better than the other (y) provided the rank of one (x) is at-least equal or same to the other (y) and the crowding distance of x is more than y.

### 3.2. NSGA-II: elitist non-dominated sorting genetic algorithm

NSGA-II [34] is also a popular multi-objective genetic algorithm based constrained optimization technique which is aimed at improving the Pareto optimal solutions using evolutionary operators such as: selection, genetic mutation and crossover.

The difference between NSGA-II and ENORA is how the calculation of the rank of the individuals in the population is performed. In ENORA, the rank of an individual in a population is the non-domination level of the individual in its slot, whereas in NSGA-II the rank of an individual in a population is the non-domination level of the individual in all the population.

## 4. Average one dependence estimator (AODE)

Naive Bayes is one of the most popular machine learning techniques which is highly efficient and computationally intensive for small dataset. This is mainly due to its attribute independence assumption based on frequency estimates that provides an accurate classifier. One variant of such a popular Naive Bayes technique is Averaged one-dependence estimator (AODE) Naive Bayes [35]. It is simple yet faster for its more simplicity in attribute independence assumption than Naive Bayes and accurate, that makes it a good competitor in the machine learning approaches.

It is worth noting here that if the attribute independence criteria are violated for any means, then the classification accuracy will be affected to a great extent. It is also observed from various research that violation is done in many cases, but if this is acceptable until the probability estimate of the most probable class outweighs the others.

During training stage, AODE produces 3-D joint frequency table with one by class values and the rest two are for attribute values. Care should be taken to ensure that missing class value are to be properly addressed by deleting the whole objects associated with that class. For missing attribute values, one can opt for allowing it for training the classifier with various possibilities of replacement. During the testing phase, conditional probability estimate is obtained from the joint frequency table. There is a significant difference in AODE and A0DE(averaged zero dependence estimators)in that while the former finds the average of all models considering direct attribute independencies, uses alternative attribute inter-dependency criteria for model aggregation, hence successfully overcomes the attribute independence problem; the averaging of aggregates models are obtained through feature subset selection in the later, hence may be trapped in attribute Independence problem. The similarity of AODE and A0DE lies with prediction by using probability estimates of several models.

**Computational complexity**

AODE may be efficient ones while applying in large dataset for its less computational complexity in terms of time and memory space, required during training and testing the classifier.

Training time complexity is $O(nk^2)$ which is linear with respect to training dataset, where n is the number of attributes and k is the number of training objects. At the same time, the time complexity for the testing an object (k) is $O(mn^2)$, where m is the number of class labels.

AODE takes very less memory space for storing the joint frequency tables is $O(mp^2)$, where m and p denotes number of class labels and number of attributes respectively. While testing, no training data are kept in memory, as training of the classifier is performed on a single sequential pass only.

**Sensitivity to large numbers of attributes**

In order to check the effectiveness of AODE classifier in handling the dataset having large number of attributes, as there is always a chance of possible dilution in attribute independence assumption. Hence, we propose to use MOEA to reduce the attribute size to make AODE is a promising classifier in comparison to the others in terms of accuracy and time to build the model.

## 5. Materials and Methods

In this section, the details about the 23 dataset consisting small, medium and large in volume, variety, velocity and veracity used in this experiment are listed in Table 1. They are collected from UCI Machine Learning repository [36].

To evaluate the performance of AODE with ENORA, all the experiments are conducted by applying them to each data set using 10-fold cross validation on a dual-processor 1.7Ghz Pentium 5 CPU using Windows with 1TB HDD, 4GB RAM in Java environment.

The proposed framework for the ENORA based multi-objective evolutionary algorithm with AODE Naive Bayes classifier is shown in Figure 2.

As can be seen, the input dataset is applied to ENORA for attribute/feature subset selection process. The detail step wise operation of ENORA is presented in the box mentioned. The obtained reduced attribute sets are now applied to AODE classifier for obtaining the meaningful knowledge in taking a decision in the whole classification process.

The model is broad when it deals with large number of variables or attributes. At the same time, the model is both broad and deep, when it finds the numerous variables maintain some kind of complex relations among themselves. In this paper, we consider Poker hand, airline are considered to be big datasets; ALL-AML leukemia, DLBCL Lymphoma, Lung Harvard, Ovarian and medical datasets are taken as medium datasets and the rests are small datasets.

The performance measures for our proposed approach ENORA-AODE to its suitability in machine learning applications to big , medium and small datasets are: accuracy, 0-1 loss and root mean square error(RMSE) and time taken to build the model.

The classification accuracy sometimes is otherwise known as 0-1 loss because at any instant of testing, the prediction may be correct or false. This is calculated either from a completely separate test dataset or else it could be a cross validation data. The 0-1 loss can be interpreted as probability estimate for which some of the randomly cross validated test data are wrongly classified. This loss should be as low as possible for a good classifier. At the same time, we may consider the classifier is a better one in comparison to others, if it achieves more accuracy, low RMSE and less time to build the model.

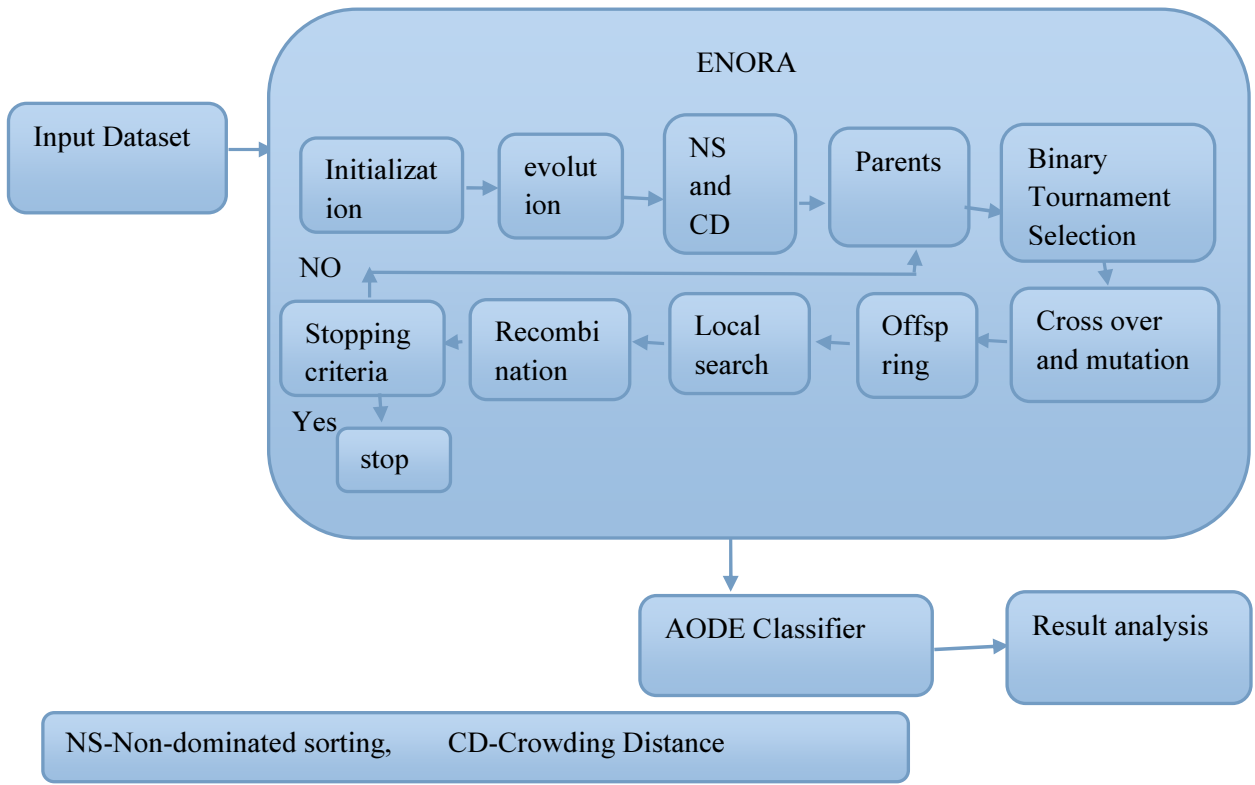

**Fig. 2: Proposed Framework**

## 6. Research Findings

In this section, the experimental results obtained are evaluated to understand the efficacy of the proposed ENORA-AODE classifier model. We perform the assessment of the classifier performance with the help of several matrices such as: accuracy, time to build the model, zero-one loss and RMSE.

At first, the experimental study is carried with 21 datasets obtained from UCI Machine learning repository using fusion of ENORA and AODE classifier. The results are presented in Table 2. It can be observed from Table 2 that for large numbers of attributes, the model building time of AODE classifier increases significantly from 0 seconds to 6.24 seconds for completing the 10-fold cross validation. For most of the dataset, it is dramatically faster takes almost no time to build the model. Their corresponding classification accuracy is good with low root mean square error (RMSE), makes the proposed approach, a competitive ones among many available machine learning algorithms. The obtained results for bio-medical datasets and two big datasets (Poker hand and Spam base) are compared with available literature in Table 3, Table 4 with accuracy and Table 5 for RMSE respectively. Further, Table 6 highlights the results with RMSE values for various real world datasets and compared with other existing works available in the literature.

**Table 1: Dataset used for experiments**

| Number | Domain | Instances | Original Attributes | Reduced attributes | Class |
|---|---|---|---|---|---|
| 1 | Colon Tumor | 62 | 2001 | 123 | binary |
| 2 | ALL-AML | 38 | 7132 | 1260 | binary |
| 3 | DLBCL | 47 | 4027 | 217 | binary |
| 4 | Lung Harvard | 203 | 12601 | 519 | multi |
| 5 | Ovarian | 253 | 15155 | 1294 | binary |
| 6 | Medical | 978 | 1494 | 316 | binary |
| 7 | Arrhythmia | 452 | 279 | 19 | multi |
| 8 | Breast Cancer | 286 | 10 | 6 | binary |
| 9 | Pima-diabetics | 768 | 9 | 5 | binary |
| 10 | Lymphography | 148 | 19 | 12 | multi |
| 11 | Hepatitis | 155 | 20 | 11 | binary |
| 12 | Heart-C | 303 | 14 | 8 | multi |
| 13 | German credit | 1000 | 21 | 4 | binary |
| 14 | Waveform | 5000 | 41 | 15 | multi |
| 15 | Nursery | 12960 | 9 | 2 | multi |
| 16 | Pen-digit | 10992 | 17 | 12 | multi |
| 17 | Poker hand | 1175067 | 10 | 6 | multi |
| 18 | Airline | 539383 | 8 | 4 | multi |
| 19 | Spam base | 4601 | 58 | 42 | binary |
| 20 | Supermarket | 4627 | 217 | 2 | binary |
| 21 | Emotion | 593 | 77 | 16 | binary |

Further for obtaining the efficiency and suitability of our proposed ENORA-AODE fusion approach, extensive comparisons are made by considering win-tie-loss criteria for the whole 22 datasets in terms of accuracy and 0-1 loss. The detailed comparison in terms of accuracy is provided in table 7 and Table 8, while for 0-1 loss interpretation, the results are provided in table 9 and Table 10.

**Table 2: Performance of our proposed approach**

| Number | Dataset | Time in seconds | Accuracy in % | RMSE |
|---|---|---|---|---|
| 1 | Colon Cancer | 0.51 | 83.87 | 0.3755 |
| 2 | ALL-AML | 0.71 | 97.37 | 0.1621 |
| 3 | DLBCL | 0.03 | 97.87 | 0.1504 |
| 4 | Lung Harvard | 0.72 | 93.11 | 0.1633 |
| 5 | Ovarian | 2.31 | 98.02 | 0.1406 |
| 6 | Medical | 0.29 | 97.03 | 0.1455 |
| 7 | Arrhythmia | 0.01 | 73.45 | 0.1616 |
| 8 | Breast Cancer | 0 | 72.73 | .4486 |
| 9 | Pima-diabetes | 0 | 75.78 | 0.4132 |
| 10 | Lymphography | 0 | 85.81 | 0.2315 |
| 11 | Hepatitis | 0 | 87.09 | 0.3165 |
| 12 | Heart-C | 0 | 83.17 | 0.2232 |
| 13 | German credit | 0 | 73 | 0.4248 |
| 14 | Waveform | 0.07 | 84.1 | 0.2742 |
| 15 | Nursery | 0.02 | 70.97 | 0.2632 |
| 16 | Pen-digit | 0.27 | 97.76 | 0.0582 |
| 17 | Poker hand | 6.24 | 71.53 | 0.2012 |

| 18 | Airline | 2.63 | 26.41 | 0.2175 |
| 19 | Spam base | 0.08 | 93.37 | 0.2381 |
| 20 | Supermarket | 0.01 | 64.03 | 0.4502 |
| 21 | Emotion | 0.01 | 84.32 | 0.3546 |

**Table 3: Performance Comparison with others work (Medical dataset) in terms of % accuracy**

| Algorithm/ Dataset | Colon Tumor | ALL-AML | DLBCL | Medical | Breast Cancer | Pima-Diabetes |
|---|---|---|---|---|---|---|
| CHCGA+SVM [16] | --- | ---- | ---- | 73.2 | 97.36 | 80.47 |
| CHCGA+RBF [16] | --- | ---- | ---- | 70 | 96.77 | 76.82 |
| K-means [21] | 78.12 | -- | 57.89 | ---- | ---- | --- |
| MOGA-3 Obj [21] | 89.58 | -- | 96.05 | ---- | ---- | --- |
| Liu and Eba [37] | 80 | -- | 90 | ---- | ---- | --- |
| Hermandev [38] | 84.6 | -- | 93.3 | ---- | ---- | --- |
| Hybrid GA-BPN [41] | --- | 89.33 | 85.65 | ---- | ---- | --- |
| Ours (ENORA-AODE) | 83.87 | 97.37 | 97.88 | 97.04 | 72.73 | 75.78 |

From Table 3, it can be observed that ours approach performs best in ALL-AML, DLBCL and medical datasets, reasonably well in Colon tumor, Pima-diabetes datasets but poor in breast cancer datasets.

**Table 4: performance measures in terms of Accuracy**

| Methodology | Accuracy in % |
|---|---|
| **German Credit** | |
| ANFIS [43] | 70 |
| MOPSO [43] | 70 |
| NSGA-II+SFP [45] | 72.9 |
| Ours (ENORA-AODE) | 73 |
| **Spam base** | |
| SSMA [44] | 91.23 |
| NSGA-II+SFP [45] | 86.79 |
| Ours (ENORA-AODE) | 93.37 |
| **Heart-C** | |
| SSMA [44] | 63.51 |
| Ours (ENORA-AODE) | 83.17 |
| **Lymphography** | |
| SSMA [44] | 55.78 |
| Ours (ENORA-AODE) | 85.81 |
| **Pima- Diabetes** | |
| SSMA [44] | 82.15 |
| NSGA-II+SFP [45] | 76.43 |
| Ours (ENORA-AODE) | 75.78 |
| **Breast Cancer** | |
| SSMA [44] | 97.65 |
| Ours (ENORA-AODE) | 72.73 |

| Nursery | |
|---|---|
| SSMA [44] | 88.07 |
| Ours (ENORA-AODE) | 70.97 |
| DLBCL | |
| Multi-objective binary PSO [46] | 91.84 |
| Ours (ENORA-AODE) | 97.87 |
| ALL-AML | |
| Multi-objective binary PSO [46] | 79.09 |
| Ours (ENORA-AODE) | 97.37 |

Table 4 confirms that our proposed approach wins the race for German credit, Spam base, Heart-C, Lymphography, DLBCL and ALL-AML datasets and loses in Pima-diabetes, Nursery and Breast cancer datasets. The majority of wins goes in favor of our proposed approach as an efficient ones among many presented in the table 4.

**Table 5: RMSE Comparison with Big dataset**

| Algorithm | Poker Hand |
|---|---|
| A1DE[40] | 0.2217 |
| A2DE [40] | 0.2044 |
| DBL1 [40] | 0.2382 |
| A1JE [40] | 0.2382 |
| Ours (ENORA-AODE) | 0.2012 |

While comparing big datasets such as poker hand, our approach concedes lowest RMSE, opines it to be the best model in comparison to other variants of AODE classifier available. It is also found that ours ENORA-AODE is the best for a relatively small datasets like: spam base with lowest RMSE.

**Table 6: RMSE comparison with real dataset**

| method | Lymphography | German Credit | Waveform | Nursery | Pen-digit | Heart-c | Emotions |
|---|---|---|---|---|---|---|---|
| NB [39] | 0.2601 | 0.4204 | 0.3413 | 0.1770 | 0.1652 | 0.3714 | -- |
| AODE [39] | 0.2542 | 0.4185 | 0.3058 | 0.1583 | 0.1001 | 0.3679 | -- |
| A2DE [39] | 0.2557 | 0.4223 | 0.3022 | 0.1428 | 0.0964 | 0.3683 | -- |
| PA2DE [39] | 0.2421 | 0.4128 | 0.3072 | 0.1428 | 0.0964 | 0.3683 | -- |
| RS [42] | --- | --- | ---- | ---- | ---- | ---- | 0.370 |
| Ours (ENORA-AODE) | 0.2315 | 0.4248 | 0.2742 | 0.2632 | 0.0582 | 0.2232 | 0.3546 |

As can be seen from Table 6 that our approach outweighs Naive Bayes, AODE, A2DE, PA2DE in Lymphography (Lymph), waveform, Pen-digit and Heart-C datasets and RS in emotions dataset; but fails in German credit and nursery datasets. Here also, majority wins supports our approach.

**Table 7: Performance in terms of accuracy**

| Dataset | Ours (ENORA-AODE) | AODE [47] | AWAODE-CFS [47] | AW-AODE-GW [47] | WAODE [47] | DT-WAODE [47] |
|---|---|---|---|---|---|---|
| Breast cancer | 72.73 | 72.53 | 72.01 | 72.26 | 71.91 | 71.63 |
| German credit | 73 | 76.42 | 76.78 | 74.75 | 76.38 | 76.26 |
| Diabetes | 75.78 | 76.37 | 75.78 | 75.36 | 75.83 | 76.45 |
| Heart-c | 83.17 | 82.48 | 82.84 | 83.83 | 82.61 | 81.92 |
| Hepatitis | 87.09 | 84.92 | 84.67 | 83.43 | 84.14 | 83.28 |

| | | | | | |
|---|---|---|---|---|---|
| Lymph | 85.81 | 86.25 | 81.52 | 80.76 | 84.16 | 81.87 |
| Waveform-5000 | 84.1 | 84.87 | 83.64 | 86.31 | 83.85 | 84.4 |

**Table 8: W-T-L comparison as a summary of experimental results**

| | AODE | AWAODE-CFS | AW-AODE-GW | WAODE | DT-WAODE | Ours (ENORA-AODE) |
|---|---|---|---|---|---|---|
| AWAODE-CFS | 0-4-3 | --- | 3-2-2 | 1-5-1 | 3-2-2 | 1-2-4 |
| AWAODE-GW | 0-1-6 | 2-2-3 | --- | 3-1-3 | 3-3-1 | 2-3-2 |
| DT-WAODE | 0-3-4 | 2-2-3 | 1-3-3 | ----- | 2-2-3 | 2-1-4 |
| Ours (ENORA-AODE) | 2-2-3 | 4-2-1 | 2-3-2 | 4-2-1 | 4-1-2 | --- |
| WAODE | 0-3-4 | 1-5-1 | 1-3-3 | --- | 3-2-2 | 1-1-5 |
| AODE | --- | 3-4-0 | 6-1-0 | 4-3-0 | 4-3-0 | 3-2-2 |

To check, whether ours ENORA-AODE approach performs better in comparison to the latest developments in variants of AODE classifiers, a comparison is provided in Table 7 and table 8 in terms if accuracy and 0-1 loss respectively.

From Table-8, we can re-confirm that our proposed approach performs well in accuracy comparison to AWAODE-CFS, WAODE and DT-WAODE with more wins for the datasets in Table 7.

Similarly, 0-1 loss criteria are used to evaluate our method with win-tie-loss (W-T-L) and variants of AODE for the datasets shown in Table 9 and Table 10. The results are encouraging as the proposed ENORA-AODE approach has won all the cases with more number of wins for all datasets.

**Table 9:** 0-1 loss comparison as a summary of experimental results

| Dataset | NB [39] | AODE [39] | A2DE [39] | PA2DE [39] | FA2DE [39] | TAN [39] | Ours (ENORA-AODE) |
|---|---|---|---|---|---|---|---|
| Lymph | 0.1696 | 0.1646 | 0.1643 | 0.1574 | 0.1624 | 0.2297 | 0.1419 |
| Hepatitis | 0.1605 | 0.1588 | 0.1604 | 0.1609 | 0.1605 | 0.1630 | 0.1291 |
| Heart-c | 0.1699 | 0.1769 | 0.1849 | 0.1833 | 0.1826 | 0.2032 | 0.1683 |
| Pima-Diabetes | 0.2752 | 0.2754 | 0.2784 | 0.2772 | 0.2772 | 0.2765 | 0.2422 |
| German-c | 0.2581 | 0.2550 | 0.2573 | 0.2542 | 0.2542 | 0.2809 | 0.27 |
| Spam base | 0.3373 | 0.3373 | 0.3373 | 0.3372 | 0.3373 | 0.3384 | 0.0663 |
| Waveform-5000 | 0.2203 | 0.1984 | 0.1955 | 0.2020 | 0.2040 | 0.2155 | 0.159 |
| Pen digits | 0.1641 | 0.0665 | 0.0534 | 0.0570 | 0.0572 | 0.0940 | 0.0224 |
| Nursery | 0.0979 | 0.0743 | 0.0546 | 0.0558 | 0.0544 | 0.0692 | 0.2903 |

**Table 10:** W/D/L comparison for decision making process

|  | NB | AODE | A2DE | PA2DE | FA2DE | TAN | Ours (ENORA-AODE) |
|---|---|---|---|---|---|---|---|
| NB | --- | 1/4/4 | 1/5/3 | 1/4/4 | 1/5/3 | 2/4/3 | 1/1/7 |
| AODE | 4/4/1 | -- | 3/4/3 | 2/4/3 | 3/4/2 | 4/3/2 | 2/0/7 |
| A2DE | 3/5/1 | 2/4/3 | --- | 1/7/1 | 1/8/0 | 6/3/0 | 2/0/7 |
| PA2DE | 4/4/1 | 3/4/2 | 1/7/1 | --- | 1/8/0 | 6/3/0 | 2/0/7 |
| FA2DE | 3/5/1 | 2/4/3 | 0/8/1 | 0/8/1 | --- | 6/3/0 | 2/0/7 |
| Ours (ENORA-AODE) | 7/1/1 | 7/0/2 | 7/0/2 | 7/0/2 | 7/0/2 | 8/0/1 | ---- |
| TAN | 3/4/2 | 2/3/4 | 0/3/6 | 0/3/6 | 0/3/6 | --- | 1/0/8 |

## 7. Conclusions

This paper introduces the fusion of AODE classifier with multi-objective evolutionary algorithm ENORA to address the large number of attributes in the dataset that sometimes not advisable for better machine learning applications. AODE for its nature of behaving linearly for the training data is computationally efficient taking less time to build the model even though the numbers of attributes are large. ENORA is a promising algorithm for its best of its capability to reduce the whole set of attributes to the best possible ones. The results obtained by using the fusion of ENORA-AODE have outperformed all the available AODE variants in terms of having better predictive accuracy, low 0-1 loss and low RMSE. In future, more investigation are proposed for deep broad learning for big data with new nature inspired algorithms, in order to obtain more effective and efficient solutions.